\documentclass[pdflatex,sn-nature]{sn-jnl}

\usepackage{graphicx}%
\usepackage{multirow}%
\usepackage{amsmath,amssymb,amsfonts}%
\usepackage{amsthm}%
\usepackage{mathrsfs}%
\usepackage[title]{appendix}%
\usepackage{xcolor}%
\usepackage{textcomp}%
\usepackage{manyfoot}%
\usepackage{booktabs}%
\usepackage{algorithm}%
\usepackage{algorithmicx}%
\usepackage{algpseudocode}%
\usepackage{listings}%

\theoremstyle{thmstyleone}%
\theoremstyle{thmstyletwo}%

\theoremstyle{thmstylethree}%

\newcommand{\ph}[1]{\texttt{[#1]}}
\renewcommand{\ph}[1]{#1}
\newcommand{\ours}{\textsc{Continuum-LM}}

\DeclareUnicodeCharacter{E5F8}{-}
\begin{document}

\title[Dynamic System Inspired Continuous Large Language Model for AI4Science Tasks]{
Dynamic System Inspired Continuous Large Language Model for AI4Science Tasks
}

\author[1,2]{\fnm{Zhaoxin} \sur{Fan}}\email{zhaoxinf@buaa.edu.cn}




\author[3]{\fnm{Zheng} \sur{Zhu}}\email{zhengzhu@ieee.org}

\author[1,2]{\fnm{Yuqing} \sur{Ma}}\email{mayuqing@buaa.edu.cn}

\author[1,2]{\fnm{Faguo} \sur{Wu}}\email{faguo@buaa.edu.cn}


\author*[1,2]{\fnm{Jian} \sur{Yang}}\email{jianyang0227@gmail.com}

\author*[1,2]{\fnm{Wenjun} \sur{Wu}}\email{wwj09315@buaa.edu.cn}


\affil[1]{\orgdiv{Beijing Advanced Innovation Center
for Future Blockchain and Privacy Computing}, \orgname{Beihang University}, \orgaddress{ \city{Beijing}, \postcode{100191}, \state{Beijing}, \country{China}}}

\affil[2]{\orgdiv{School of Artificial Intelligence}, \orgname{Beihang University}, \orgaddress{\city{Beijing}, \postcode{100191}, \state{Beijing}, \country{China}}}

\affil[3]{\orgname{GigaAI}, \orgaddress{\city{Beijing}, \postcode{100191}, \state{Beijing}, \country{China}}}

\abstract{Scientific foundation models increasingly operate on continuous phenomena, yet still use discrete residual computation.
We ask whether pretrained residual transformations can support additional latent computation without expanding the backbone.
To test this possibility, we introduce Continuum-LM, which reuses these transformations as dynamical operators in a continuous-depth language model.
Its hidden states evolve through an ordinary differential equation evaluated by higher-order numerical integration, with solver order and step size controlling computation.
Across eight scientific tasks, Continuum-LM trains stably and improves mean exact-match accuracy by 7.77 and 10.96 percentage points over discrete counterparts at the \ph{0.6B} and \ph{1.7B} scales, respectively.
At \ph{0.6B}, it also outperforms an Ouro-style recurrent control under matched parameter and residual-function evaluation budgets.
Together, these results support extensible computational depth through parameter reuse, identifying continuous hidden-state evolution as a practical design dimension for scientific language models beyond parameter scaling.}


\maketitle

\section{Introduction}

Artificial intelligence is increasingly used to model, predict and design scientific systems, with foundation models contributing to molecular and protein modelling, materials science, biomedical reasoning and scientific problem-solving~\cite{jumper2021highly,abramson2024accurate,edwards2022molt5,gu2021pubmedbert,bran2024chemcrow}. Large language models (LLMs) provide a common backbone for these tasks through domain-specific pretraining or supervised fine-tuning~\cite{Beltagy2019scibert,luo2022biogpt,taylor2022galactica,lei2024materialsllm}. Although these approaches have broadened the scope of scientific foundation models, they have largely expanded training data, domain knowledge and parameter count while retaining the fixed-depth architecture of general-purpose LLMs.

Across these scientific tasks, predictions must satisfy numerical, symbolic or structural constraints despite scarce labels and differences between training and deployment systems~\cite{madani2023progen,zhang2024chemllm,lei2024materialsllm}. Meeting these demands depends not only on the scientific information encoded in a model, but also on how its internal representations are transformed during computation. This motivates an architectural question that has received comparatively limited attention in scientific foundation models: must hidden-state evolution follow a fixed sequence of residual updates, or can computation be controlled through integration depth and numerical resolution?

This question can be formulated through the residual update
\begin{equation}
    h_{j+1}=h_j+F_j(h_j),
    \label{eq:discrete_block}
\end{equation}
where $h_j$ is the hidden state entering residual sublayer $j$
and $F_j$ is its residual
transformation~\cite{vaswani2017attention,he2016deep}.
Under an appropriate parameterization, this update can be interpreted as a numerical discretization of continuous-depth dynamics~\cite{weinan2017proposal,haber2018stable,chen2018neural,ruthotto2020deep}. Rather than implying that discrete networks are intrinsically inaccurate or must mirror physical continuity, this interpretation makes solver order and integration resolution explicit choices for regulating numerical approximation and computational cost within each branch.

Applying this interpretation to pretrained LLMs introduces two requirements. First, conversion must reuse existing parameters and support stable optimization rather than require training from scratch. Second, its organization of additional function evaluations must be compared with another form of repeated computation at the same parameter scale. The challenge is therefore to adapt an already capable discrete model's computation while retaining its pretrained initialization.

\begin{figure*}[h]
\centering
\includegraphics[width=1.\textwidth]{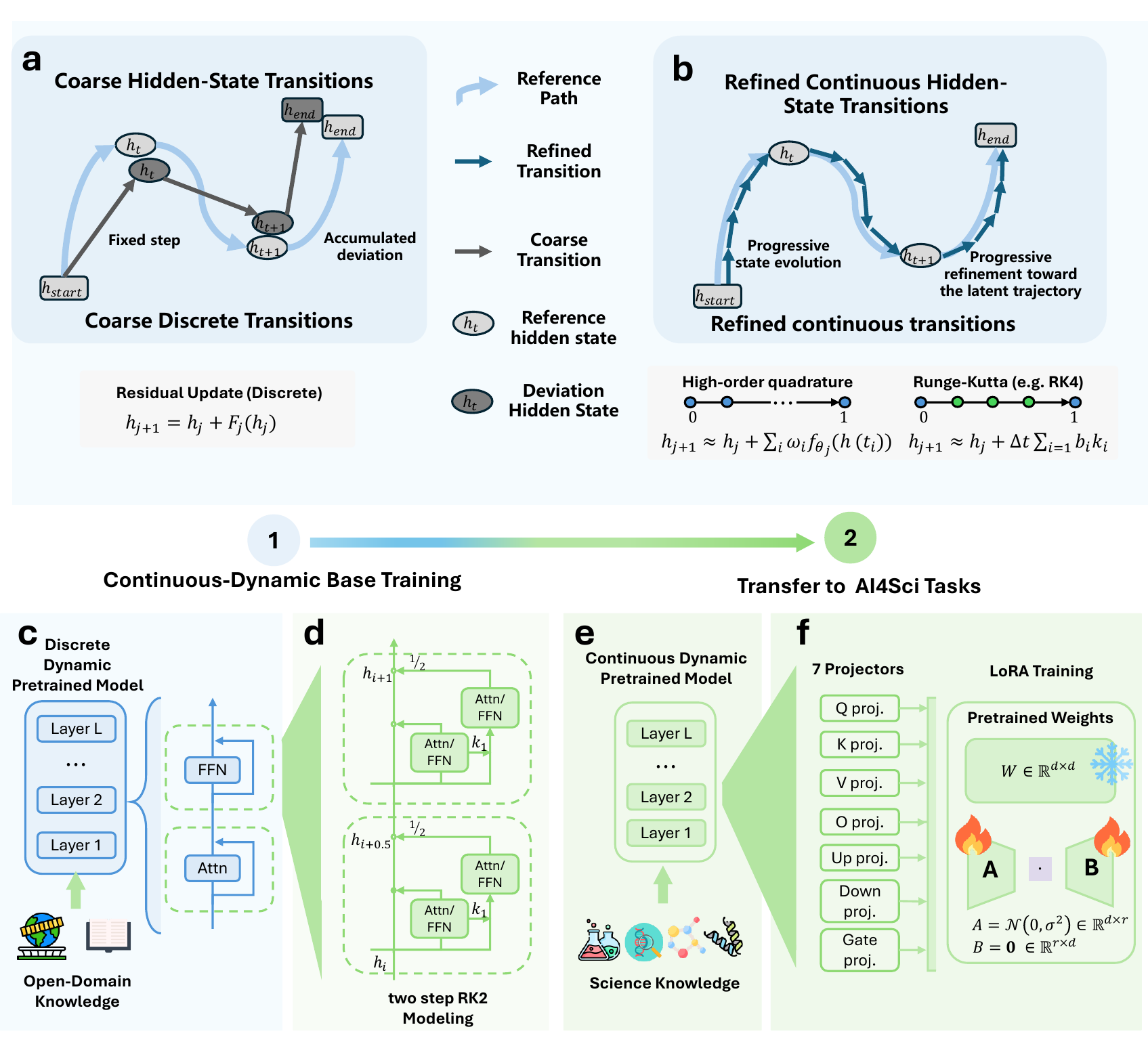}
\caption{\textbf{Continuous-depth conversion of pretrained language models.}
\textbf{a,} A standard residual network propagates hidden representations through discrete layer-wise updates. The smooth curve is an illustrative continuous-depth path rather than an assumed ground-truth trajectory; coarse discretization can produce local endpoint discrepancies that accumulate with depth.
\textbf{b,} Numerical solvers evaluate the learned update function at intermediate hidden states within the same normalized depth interval. Higher-order quadrature and Runge--Kutta (RK) methods are shown schematically as representative integration strategies.
\textbf{c,} Starting from a pretrained discrete language model, we replace each residual update with a local continuous-depth initial-value problem initialized from the corresponding pretrained parameters, followed by continued pretraining to adapt the model to the converted computation.
\textbf{d,} Within each Transformer block, the self-attention and feed-forward network (FFN) residual branches are converted separately and integrated sequentially using the default two-step midpoint RK2 solver. Solver stages share the parameters of the corresponding pretrained branch but recompute its normalization and transformation at each stage-specific hidden state.
\textbf{e,} The converted model is subsequently adapted to eight AI4Science tasks using low-rank adaptation (LoRA).
\textbf{f,} LoRA adapters are inserted into the query, key, value and output projections of self-attention and the up, gate and down projections of the FFN, while the continuous-depth backbone is frozen.}

\label{fig:framework}
\end{figure*}

To investigate this possibility, we introduce \ours{}, which parameterizes hidden-state evolution as an autonomous ordinary differential equation (ODE),
\begin{equation}
    \frac{\mathrm{d}h(t)}{\mathrm{d}t}=f_{\theta}(h(t)),
    \label{eq:continuous_hidden_dynamics}
\end{equation}
whose update function inherits parameters from a discrete checkpoint and is evaluated by numerical integration (Fig.~\ref{fig:framework}). Continuous-depth base training first adapts these parameters to the modified computation. Low-rank adaptation (LoRA) then specializes the converted model to scientific tasks while keeping the backbone frozen~\cite{hu2022lora}. The two stages thus separate adaptation of pretrained computation from downstream scientific specialization.

We evaluate this framework through parameter-scale suitability, optimization and scientific task adaptation. Layer-wise sweeps identify minimum-loss coefficients near the original pretrained scale, while solver comparisons reveal integration-dependent training dynamics (Fig.~\ref{fig:base_validation}a--c). After downstream adaptation, \ours{} improves mean exact-match accuracy by 77.2\% and 82.0\% over corresponding discrete baselines at \ph{0.6B} and \ph{1.7B}, respectively; the \ph{1.7B} model approaches Qwen3-\ph{8B} (Fig.~\ref{fig:ai4sci_results}). To compare alternative organizations of repeated computation, we also evaluate a matched recurrent control at \ph{0.6B} (Fig.~\ref{fig:compute_efficiency}e).

Together, these results identify continuous depth as a practical architectural variable rather than an analogy for a uniquely defined reasoning trajectory. By linking parameter reuse to numerical integration and downstream adaptation, they suggest that scientific language models can benefit from changes to how pretrained representations are computed, alongside changes in data, parameters and fine-tuning.

\section{Results}\label{sec:results}

\subsection{Pretrained residual LLMs admit direct continuous-depth conversion}
\label{subsec:overview}

To implement this continuous-depth view, we first transferred pretrained residual computation without reinitializing or reshaping the weights. For the branch in Eq.~\ref{eq:discrete_block}, we define the local system
\begin{equation}
    \frac{\mathrm{d}h(t)}{\mathrm{d}t}
    =F_j(h(t)),
    \qquad h(0)=h_j,\qquad t\in[0,1].
    \label{eq:continuous_block}
\end{equation}
Under this formulation, the original update is a unit-step Euler approximation, whereas higher-order integration evaluates the inherited transformation at intermediate states~\cite{chen2018neural,butcher2016numerical}. The numerical update can therefore change without adding backbone parameters. We examine a two-stage second-order Runge--Kutta method (RK2) and a four-stage fourth-order method (RK4), varying both numerical order and integration resolution.

This local construction is applied independently to the attention and feed-forward residual sublayers (Fig.~\ref{fig:framework}a--d). Their pretrained parameters define the respective vector fields, while the tokenizer, embeddings, normalization structure, positional encoding and output head retain their original forms. A discrete checkpoint can therefore initialize the converted model without reinitialization or format changes. The conversion itself supplies a valid initialization before any gradient-based adaptation. Subsequent continuous-depth base training adapts these inherited parameters to the numerical update. Backbone parameter count remains unchanged, whereas function evaluations and computational cost depend on the solver.

By retaining the pretrained initialization and high-level structure while changing the numerical update, this construction makes within-block computation an explicit architectural intervention. It allows multi-stage residual evaluation to be compared with the original single-stage update without increasing parameter count, rather than assuming access to a unique underlying reasoning trajectory. Solver order and evaluation count can consequently be varied independently of model parameter count.

We assessed the inherited parameter scale using Qwen3-\ph{0.6B} and Qwen3-\ph{1.7B}~\cite{yang2025qwen3}. For each layer--branch configuration, we rescaled the selected layer's pretrained parameters, kept all others unchanged, and measured teacher-forced next-token cross-entropy on 1,000 sampled training examples. Minimum-loss coefficients clustered near $c=1$ across most attention and feed-forward configurations (Fig.~\ref{fig:base_validation}a), supporting the original parameter scale as an initialization for continued training.

Starting from these checkpoints, all evaluated RK2 and RK4 configurations showed an overall loss reduction during 20,000-step solver ablations (Fig.~\ref{fig:base_validation}b,c). The main models use a separate 100,000-step continued-pretraining budget. This distinction separates the diagnostic training curves from the checkpoints used for downstream evaluation.
We then apply task-specific LoRA fine-tuning while freezing the converted backbone (Fig.~\ref{fig:framework}e,f)~\cite{hu2022lora}. The resulting protocol evaluates the same continuous-depth base model across scientific domains without independently retraining its full backbone for each task.


\begin{figure*}[t]
\centering
\includegraphics[width=1.\textwidth]{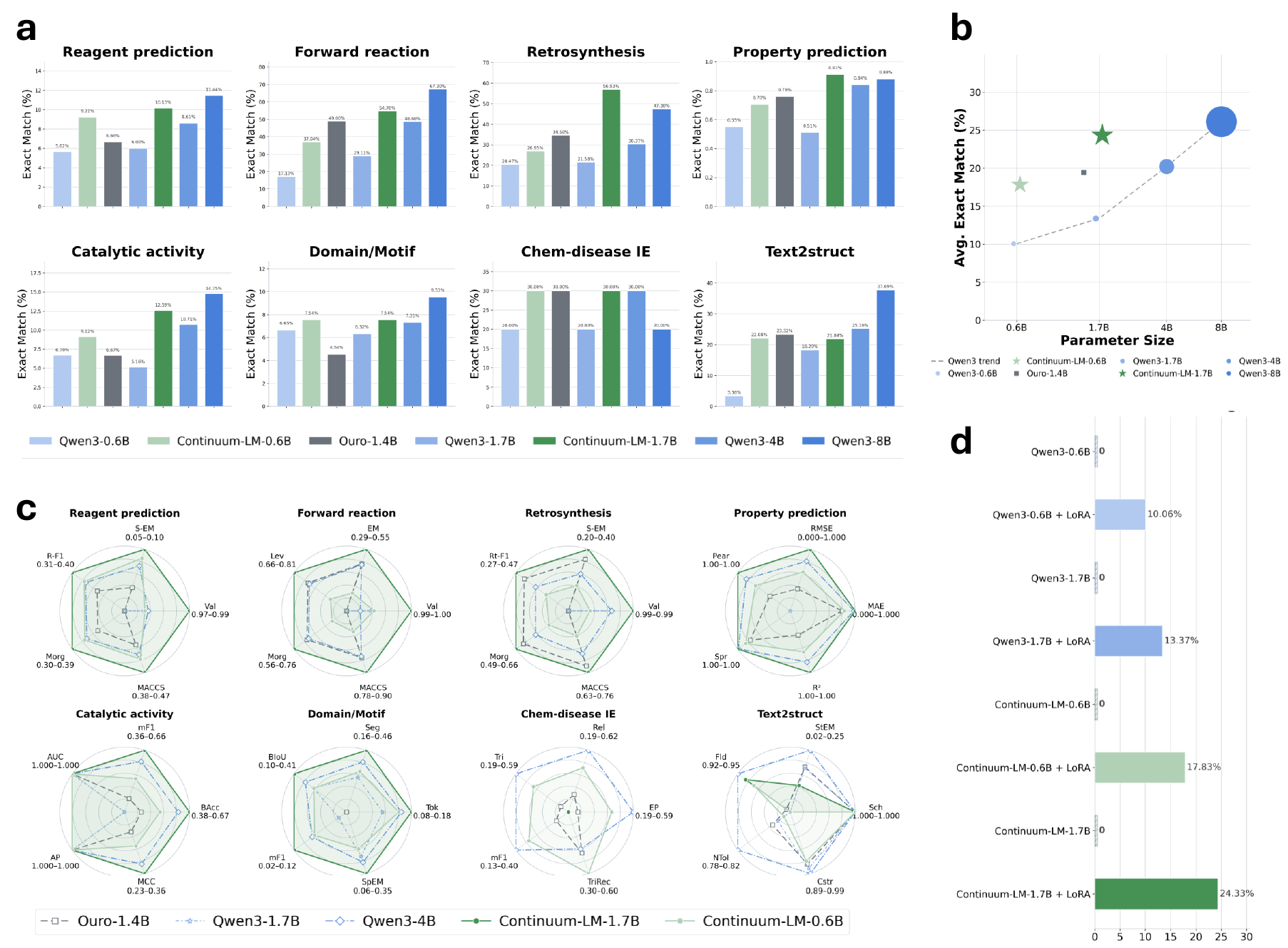}
\caption{\textbf{Performance across eight AI4Science tasks.}
\textbf{a,} Exact-match accuracy of \ours{} and baseline models across eight tasks spanning four scientific task families.
\textbf{b,} Parameter--performance trade-off, measured by mean exact-match accuracy across the eight tasks. Circles indicate supervised fine-tuned (SFT) Qwen3 baselines, the square indicates Ouro-\ph{1.4B}, and stars indicate \ours{} models; the dashed line represents the empirical scaling trend fitted to the Qwen3 SFT results. \ours{}-\ph{0.6B} outperforms Qwen3-\ph{1.7B}@SFT, whereas \ours{}-\ph{1.7B} outperforms Qwen3-\ph{4B}@SFT and achieves performance comparable to that of Qwen3-\ph{8B}@SFT.
\textbf{c,} Task-specific evaluation across five complementary metrics per task. Each axis is direction-aligned and min--max scaled across the models displayed in the panel. 
\textbf{d,} Mean exact-match accuracy before and after LoRA fine-tuning for \ours{} and parameter-matched Qwen3 models. Post-LoRA scores are averaged over three independent fine-tuning runs from each fixed base checkpoint.
}
\label{fig:ai4sci_results}
\end{figure*}

\subsection{Continuous-depth models improve performance across eight scientific tasks}
\label{subsec:main_results}

To evaluate this adaptation protocol across scientific domains, we used eight tasks spanning four families. This coverage assesses the converted backbone beyond a single benchmark or data modality. Molecular tasks comprise reagent prediction, forward reaction prediction, retrosynthesis and property prediction~\cite{andronov2023reagent,schwaller2019molecular,yang2019molecular,wu2018moleculenet}; protein tasks comprise catalytic activity and domain/motif prediction~\cite{ryu2019deepec,yu2023enzyme,mistry2021pfam,blum2025interpro}. The remaining tasks are chemical--disease relation extraction~\cite{li2016bc5cdr,wei2016biocreativecdr} and materials text-to-structure generation~\cite{flamshepherd2023language,antunes2024crystallm,jain2013materialsproject}. At each scale, the continuous model starts from the corresponding Qwen3 checkpoint. Downstream comparisons share task data, LoRA rank, optimization schedules and evaluation procedures; scores average three independent LoRA runs from each fixed base checkpoint.

Under this protocol, \ours{} outperformed its parameter-matched Qwen3 counterpart on all eight tasks at both scales (Fig.~\ref{fig:ai4sci_results}a). Forward reaction accuracy increased from 17.13\% to 37.04\% at \ph{0.6B} and from 29.11\% to 54.70\% at \ph{1.7B}, gains of 19.91 and 25.59 percentage points. At \ph{0.6B}, text-to-structure accuracy rose from 3.36\% to 22.08\%, a gain of 18.72 points. Improvements also occurred in retrosynthesis, catalytic activity and chemical--disease interaction extraction. Across the complete benchmark suite, gains ranged from 0.154 points for \ph{0.6B} property prediction to 35.35 points for \ph{1.7B} retrosynthesis. Thus, the gains spanned the task suite, with the largest absolute improvements in structured scientific outputs.

To relate these task-level gains to model size, we computed the unweighted mean exact-match accuracy across the eight tasks (Fig.~\ref{fig:ai4sci_results}b). \ours{}-\ph{0.6B} achieved 17.83\%, compared with 10.06\% for Qwen3-\ph{0.6B}, 13.37\% for Qwen3-\ph{1.7B} and 20.21\% for Qwen3-\ph{4B}. At \ph{1.7B}, \ours{} reached 24.33\%, exceeding Qwen3-\ph{4B} by 4.12 percentage points and approaching Qwen3-\ph{8B} (26.12\%) within 1.79 points. These comparisons show that the parameter--performance advantage extends beyond same-scale baselines: the continuous \ph{0.6B} and \ph{1.7B} models also outperform the larger discrete \ph{1.7B} and \ph{4B} models, respectively.

We next examined whether the improvements extended beyond a single definition of task success. The radar analysis covers 40 task--metric combinations, with five metrics per task and min--max scaling across the displayed models (Fig.~\ref{fig:ai4sci_results}c). These radial values describe relative performance within each metric, rather than replacing the raw task scores used in the aggregate comparison. Molecular gains extend to fingerprint similarity as well as exact matching, while protein improvements persist under class-sensitive and span-level criteria. Text-to-structure generation remains competitive in schema validity, field-level F1 and numerical-tolerance accuracy. These complementary measures support broader output-quality improvements, although their magnitude varies by task and metric.

To examine the role of downstream adaptation, we compared each base model before and after LoRA fine-tuning (Fig.~\ref{fig:ai4sci_results}d). Both discrete and continuous checkpoints initially achieved near-zero aggregate exact match. With matched trainable parameter counts and optimization settings, their post-LoRA scores reached 10.06\% versus 17.83\% at \ph{0.6B} and 13.37\% versus 24.33\% at \ph{1.7B}, respectively. The observed differences therefore concern how the two types of base checkpoint respond to the same task-specific training procedure. The continuous-dynamics base models support stronger downstream solutions under the same parameter-efficient adaptation protocol.

\begin{figure}
    \centering
    \includegraphics[width=0.9\linewidth]{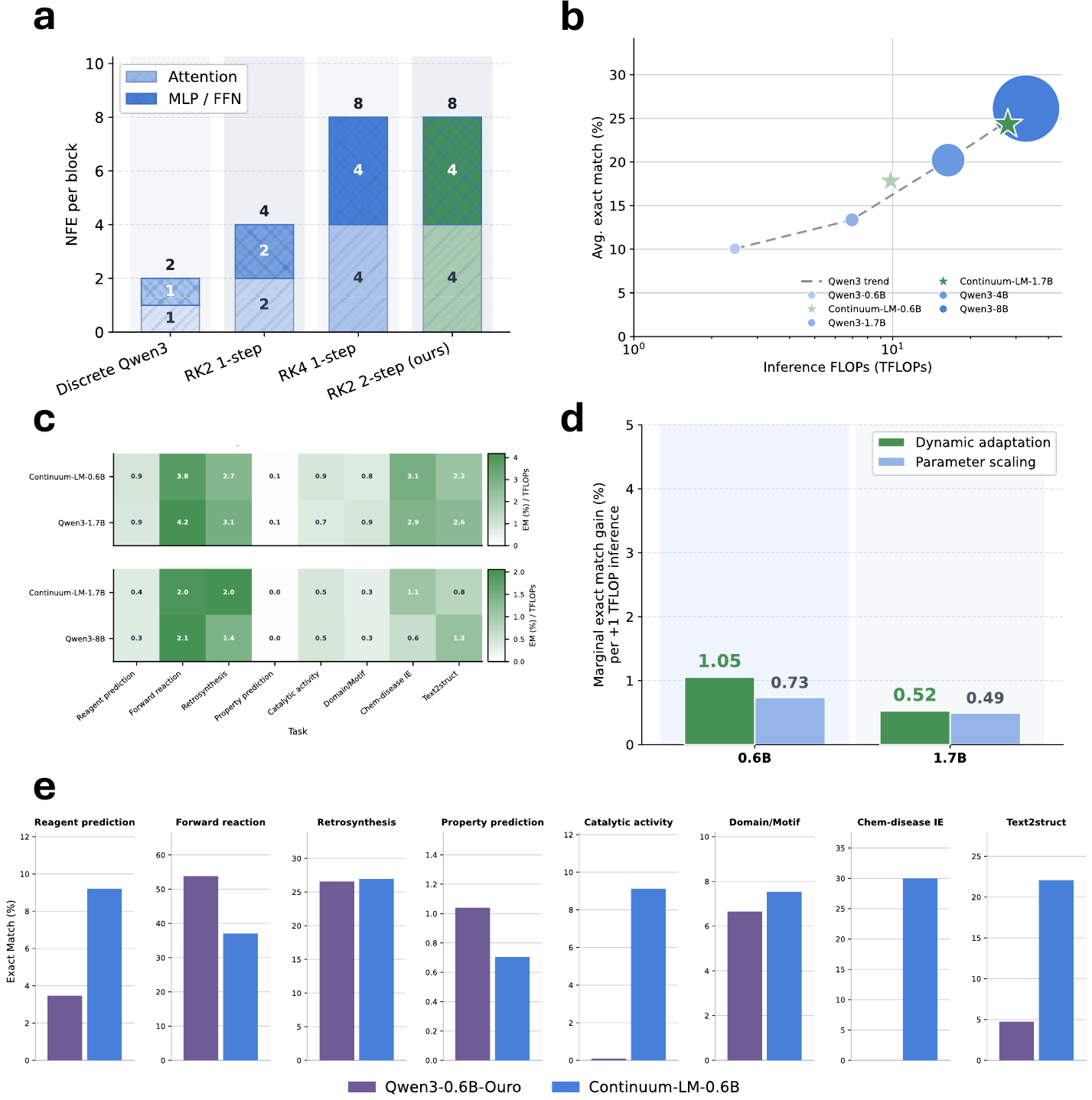}
    \caption{\textbf{Inference-compute trade-offs and recurrent-control comparisons.}
    \textbf{a,} Numbers of function evaluations for the attention
    and feed-forward residual branches within a Transformer block.
    Each branch is evaluated once in the discrete model, twice
    with one-step RK2, and four times with either one-step RK4
    or two-step midpoint RK2. Stacked bars show the combined
    attention and feed-forward evaluation counts.
    \textbf{b,} Mean exact-match accuracy across eight AI4Science
    tasks versus estimated inference FLOPs. Circles denote Qwen3
    baselines, stars denote \ours{} models, and the dashed line
    shows the empirical Qwen3 compute--performance trend.
    \textbf{c,} Task-level exact-match accuracy per TFLOP.
    Each cell reports exact-match accuracy expressed as a percentage
    divided by estimated inference cost in TFLOPs per example.
    The upper heatmap compares \ours{}-\ph{0.6B} with
    Qwen3-\ph{1.7B}; the lower heatmap compares
    \ours{}-\ph{1.7B} with Qwen3-\ph{8B}.
    \textbf{d,} Marginal gain in mean exact-match accuracy per
    additional TFLOP. Continuous-depth conversion is compared
    with discrete scaling from \ph{0.6B} to \ph{1.7B} and
    from \ph{1.7B} to \ph{8B}, respectively.
    \textbf{e,} Task-level comparison of \ours{}-\ph{0.6B}
    and Qwen3-\ph{0.6B}-Ouro under a common pretrained
    initialization, base-training corpus and downstream adaptation
    configuration, with four function evaluations per residual
    branch. Bars show mean exact-match accuracy over three
    LoRA fine-tuning runs from each fixed base checkpoint.}
    \label{fig:compute_efficiency}
\end{figure}

Because continuous-depth conversion adds repeated function evaluations, we also compared \ours{}-\ph{0.6B} with Qwen3-\ph{0.6B}-Ouro, an Ouro-style recurrent control (Fig.~\ref{fig:compute_efficiency}e). Both share the pretrained Qwen3 checkpoint, base-training corpus and downstream configuration, with four evaluations per original attention or feed-forward residual branch. Matching these conditions compares two different ways to use repeated evaluations of pretrained transformations. Across three LoRA runs from each fixed base checkpoint, mean accuracy was 17.83\% for \ours{} and 12.04\% for the control, a 5.79-percentage-point difference; \ours{} was higher on six tasks. This comparison supports its organization of within-block computation over the evaluated recurrent alternative at matched parameter and branch-evaluation budgets.


\subsection{Inference-compute trade-offs and recurrent-control comparisons}
\label{subsec:compute_efficiency}

The recurrent comparison controls the number of residual evaluations; we next examined the inference cost associated with those evaluations. One branch-level function evaluation (NFE) is one call to an attention or feed-forward residual transformation. The discrete model uses one call per branch, one-step RK2 uses two, and one-step RK4 and two-step midpoint RK2 each use four (Fig.~\ref{fig:compute_efficiency}a). The default configuration therefore quadruples branch evaluations without changing the backbone. Because the two branch types have different costs, these counts must be complemented by FLOP estimates.

We estimated inference FLOPs per example using 1,024 input tokens,
1,024 output tokens, batch size one and key--value caching.
Using a parameter-based forward-pass approximation
motivated by the leading-order Transformer compute
estimate~\cite{kaplan2020scaling}, we calculate
\begin{equation}
    C_{\mathrm{infer},q}
    \approx
    2\nu_q P\left(T_{\mathrm{in}}+T_{\mathrm{out}}\right),
    \label{eq:inference_flop_estimate}
\end{equation}
where $P$ is the nominal parameter count in individual parameters,
$T_{\mathrm{in}}=T_{\mathrm{out}}=1{,}024$, and $\nu_q$ is the
number of evaluations per residual branch: one for the discrete
model and four for the default two-step midpoint RK2 configuration.
The NFE multiplier is applied at the model level as a leading-order
approximation, rather than through an operator-by-operator count.
Costs are divided by $10^{12}$ to obtain TFLOPs per example.
These estimates use common sequence lengths across tasks, while
performance is the unweighted mean exact-match accuracy over
the eight benchmarks.

Under this protocol, \ours{}-\ph{0.6B} achieved 17.83\% accuracy at 9.83 TFLOPs, compared with 13.37\% at 6.96 TFLOPs for Qwen3-\ph{1.7B} (Fig.~\ref{fig:compute_efficiency}b). The gain of 4.46 percentage points requires 41.2\% more computation. Compared instead with Qwen3-\ph{4B} (20.21\% at 16.38 TFLOPs), the same model uses 40.0\% less computation with a 2.38-point accuracy gap. It therefore occupies an intermediate operating point rather than dominating either baseline in both accuracy and estimated computation.

A similar trade-off appears at \ph{1.7B}: \ours{} achieved 24.33\% at 27.85 TFLOPs, compared with 26.12\% at 32.77 TFLOPs for Qwen3-\ph{8B}. It used approximately 15.0\% less inference computation with a 1.79-percentage-point accuracy gap, recovering 86\% of the performance difference between Qwen3-\ph{1.7B} and Qwen3-\ph{8B}. The continuous model thus approaches the larger model's mean performance while using fewer parameters and less estimated inference computation. Together, these operating points show competitive compute--performance trade-offs for smaller continuous models, without implying superiority at every computation budget.

To resolve these aggregate comparisons by task, we divided each task's percentage accuracy by estimated inference TFLOPs (Fig.~\ref{fig:compute_efficiency}c). \ours{}-\ph{0.6B} exceeded Qwen3-\ph{1.7B} on reagent prediction, catalytic activity and chemical--disease relation extraction. At \ph{1.7B}, \ours{} exceeded Qwen3-\ph{8B} on five tasks, most clearly in retrosynthesis and chemical--disease relation extraction; catalytic activity was nearly matched. The heatmaps thus reveal task-level differences within the aggregate compute--performance pattern.

Whereas these comparisons describe individual operating points, we also compared the return from additional inference computation relative to a shared starting model (Fig.~\ref{fig:compute_efficiency}d). For score and cost changes from $(S_0,C_0)$ to $(S_1,C_1)$, we define
\begin{equation}
    R_{\mathrm{compute}}
    =
    \frac{S_1-S_0}{C_1-C_0},
    \label{eq:marginal_compute_return}
\end{equation}
where $S$ is mean exact-match accuracy expressed as a percentage and $C$ is estimated inference cost in TFLOPs. This measures an intervention relative to its discrete source, rather than dividing its final score by total compute. Starting from Qwen3-\ph{0.6B}, conversion yielded 1.05 percentage points per additional TFLOP, compared with 0.73 for scaling to Qwen3-\ph{1.7B}. From Qwen3-\ph{1.7B}, the corresponding returns were 0.52 for conversion and 0.49 for scaling to Qwen3-\ph{8B}. Conversion provided the larger marginal return along these two paths, most clearly at \ph{0.6B}.

These comparisons characterize parameter--performance and inference-compute trade-offs under the standardized sequence-length protocol, with cost expressed as estimated FLOPs. We next turn from these operating points to the design choices governing conversion and adaptation.

\begin{figure*}[t]
\centering
\includegraphics[width=1.\textwidth]{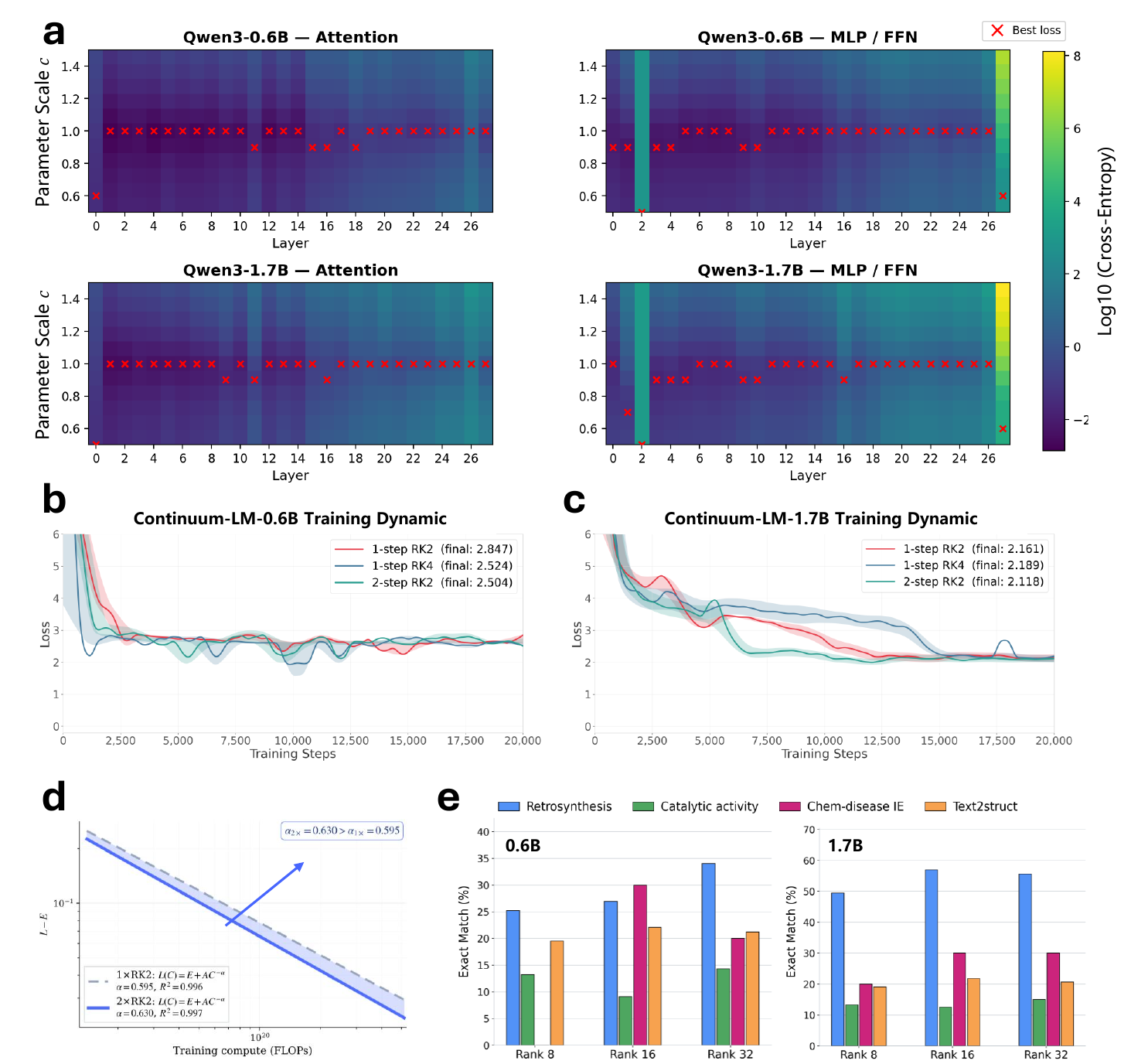}
\caption{\textbf{Ablations of continuous-depth conversion, numerical integration and downstream adaptation.}
\textbf{a,} Layer-wise parameter-scale sweep for RK2-based conversion of Qwen3-\ph{0.6B} and Qwen3-\ph{1.7B}. Attention and feed-forward residual branches are evaluated separately. Colours encode the base-10 logarithm of teacher-forced next-token cross-entropy on 1,000 randomly sampled training examples, and red crosses mark the loss-minimizing coefficient for each layer--branch configuration. Most selected coefficients lie near $c=1$, corresponding to the original pretrained parameter scale.
\textbf{b,c,} Training-loss trajectories over 20,000 optimization
steps for \ours{}-\ph{0.6B} (b) and \ours{}-\ph{1.7B} (c),
comparing one-step RK2, one-step RK4 and two-step midpoint RK2
under a common pretrained initialization and training protocol.
These solver-ablation runs use a shorter budget than the
100,000-step continued pretraining of the main models.
\textbf{d,} Power-law fits of WikiText-103 test loss against
estimated cumulative training FLOPs for the \ph{1.7B} one-step RK2
and two-step midpoint RK2 ablation runs. The vertical axis
shows excess test loss, $L-E$, where $E$ is the asymptotic
loss fitted separately for each configuration. Shading
visually distinguishes the fitted trends and does not
indicate confidence intervals.
\textbf{e,} Task-level exact-match accuracy at LoRA ranks
8, 16 and 32 for \ours{}-\ph{0.6B} (left) and
\ours{}-\ph{1.7B} (right).
Colours distinguish retrosynthesis, catalytic activity
prediction, chemical--disease relation extraction and
text-to-structure generation.
Increasing rank from 8 to 16 improves three of the four
tasks at both model scales, whereas increasing rank
further produces task-dependent gains and losses.
The rank-16 scores are the corresponding task results
used in the main benchmark comparison.
}
\label{fig:base_validation}
\end{figure*}

\subsection{Ablations identify the design factors underlying \ours}
\label{subsec:ablation}

The ablations examine four design factors: the inherited parameter scale, optimization after conversion, numerical integration and downstream LoRA capacity. We consider initialization and base training before turning to downstream adapter capacity, following the conversion and adaptation pipeline. These diagnostics characterize the conditions under which the framework is effective.

\paragraph{Pretrained parameter scales remain suitable after conversion.}
We first tested whether the original parameter scale remains suitable after RK2 conversion. For each layer--branch configuration in Qwen3-\ph{0.6B} and Qwen3-\ph{1.7B}, we rescaled only the selected layer's parameters as $\theta^{(c)}=c\,\theta$, leaving all other parameters unchanged; $c=1$ retains the pretrained values. Teacher-forced language-modelling loss was measured on 1,000 sampled training examples using the reference prefix at each position, without gradient-based parameter updates. Across most attention and feed-forward configurations, the minimum-loss coefficient remained near unity (Fig.~\ref{fig:base_validation}a). Thus, retaining the inherited parameter scale generally yielded lower prediction loss than substantial rescaling, supporting its use for initialization.

\paragraph{Integration schemes shape base-training dynamics under a common pretrained initialization.}

Starting from the same pretrained Qwen3 checkpoint at each scale, we then compared one-step RK2, one-step RK4 and two-step midpoint RK2 over 20,000 optimization steps with common data and optimizer settings (Fig.~\ref{fig:base_validation}b,c). Although loss decreased for all configurations, their trajectories differed. At \ph{0.6B}, the curves largely overlapped after the initial decrease; at \ph{1.7B}, two-step RK2 reached a lower-loss regime earlier in the middle of training. The differences narrowed later, but two-step RK2 attained the lowest reported endpoint loss at both scales. These endpoints belong to the solver ablations, not the separate 100,000-step main training, and show that the numerical update influences optimization under a common initialization.

\paragraph{Integration configurations exhibit different compute--loss scaling behaviour.}

To relate these training dynamics to computation, we analysed
the \ph{1.7B} one-step RK2 and two-step midpoint RK2 ablation runs.
For each solver configuration $q$, we fitted next-token
cross-entropy measured on the official WikiText-103 test split to
\begin{equation}
L_q(C)=E_q+A_qC^{-\alpha_q},
\label{eq:compute_loss_scaling}
\end{equation}
where $C$ denotes cumulative training compute. The asymptotic
loss $E_q$, coefficient $A_q$ and exponent $\alpha_q$ were
jointly estimated separately for each configuration.
Figure~\ref{fig:base_validation}d plots excess test loss
$L_q(C)-E_q$, abbreviated as $L-E$.

Both configurations followed a power-law trend over the measured range. One-step and two-step RK2 yielded exponents of 0.595 and 0.630, with $R^2=0.996$ and $0.997$, respectively. The larger exponent for two-step RK2 indicates a steeper decline in fitted excess test loss as compute increases. These exponents complement the step-based training curves by describing how each configuration's excess loss changes on the compute axis. Together, the two analyses support the selected solver within the evaluated regime. Both RK2 configurations remain second-order methods.

\paragraph{Moderate LoRA rank provides an effective adaptation setting.}

After examining base training, we assessed downstream adapter capacity on retrosynthesis, catalytic activity prediction, chemical--disease relation extraction and text-to-structure generation (Fig.~\ref{fig:base_validation}e). Increasing LoRA rank from 8 to 16 improved retrosynthesis, chemical--disease relation extraction and text-to-structure generation at both model scales, whereas catalytic activity prediction declined.

Increasing rank further to 32 produced mixed responses. At \ph{0.6B}, retrosynthesis and catalytic activity improved, while chemical--disease relation extraction and text-to-structure generation declined. At \ph{1.7B}, only catalytic activity improved; chemical--disease relation extraction was unchanged and the other tasks declined. Larger adapters therefore did not yield uniformly better results. We retain $r=16$ as a moderate-capacity default that balances the evaluated tasks without doubling adapter rank.

Together, these diagnostics support pretrained-scale initialization, two-step midpoint RK2 and moderate-rank LoRA as the selected configuration. Having characterized these design choices, we next examine the output-level changes accompanying the downstream gains.

\subsection{\ours{} assigns larger margins to reference scientific tokens}
\label{subsec:interpretability}

To characterize reference-token preference, we computed the teacher-forced margin for prompt $x$, reference answer $y=(y_1,\ldots,y_T)$ and answer position $t$:
\begin{equation}
    m_t=z_{\theta}(y_t\mid x,y_{<t})
    -\max_{v\ne y_t}z_{\theta}(v\mid x,y_{<t}),
    \label{eq:gt_margin}
\end{equation}
where $z_{\theta}(v\mid x,y_{<t})$ is the logit for vocabulary token $v$ under the reference prefix. A positive margin places the reference token first; a larger positive margin indicates greater separation from its strongest competitor. The maximum excludes the reference token, so the competitor is not assumed to be the second-ranked token when the reference is not ranked first. This measures conditional token preference rather than calibrated confidence.

\begin{figure*}[t]
\centering
\includegraphics[width=1.\textwidth]{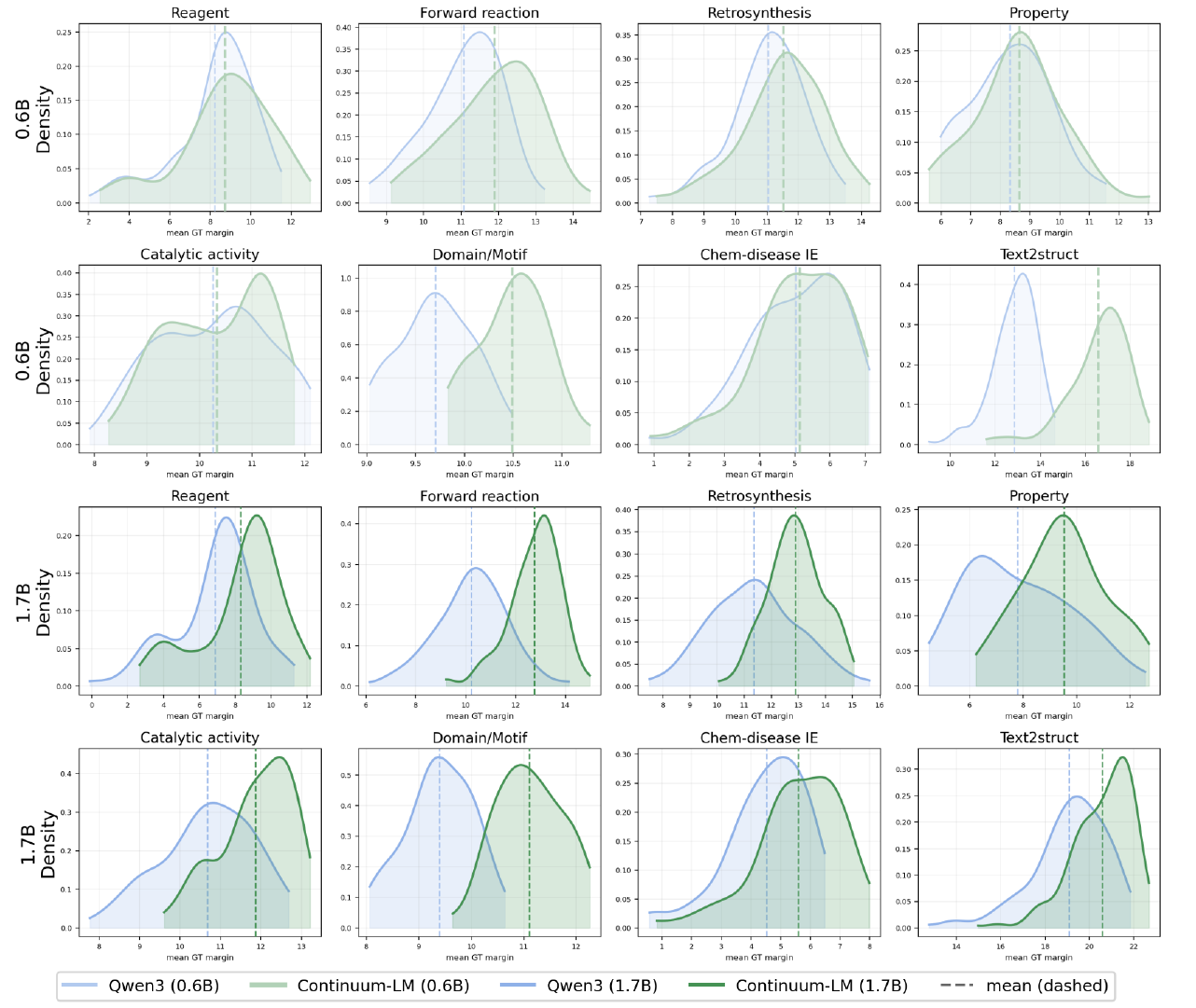}
\caption{\textbf{Token-level logit-margin analysis on AI4Science tasks.}
Distributions of reference-token logit margins on the AI4Science test sets. Under teacher-forced evaluation, the margin at each answer position is defined as the logit assigned to the reference token minus the largest logit assigned to any alternative vocabulary token. Positive values indicate that the reference token is ranked first, whereas larger values indicate stronger separation from its closest competitor. Dashed vertical lines mark the mean margin of each distribution. Relative to the corresponding Qwen3 baselines, \ours{} exhibits higher mean margins across the evaluated tasks, with clearer distributional separation
across a broader range of tasks at the \ph{1.7B} scale.}
\label{fig:gt_margin}
\end{figure*}

We compared Qwen3 and \ours{} on identical prompt--answer examples and reference prefixes at both scales. Figure~\ref{fig:gt_margin} summarizes the distributions separately for each of the eight tasks, with dashed vertical lines marking their mean margins.

At \ph{0.6B}, the distributions overlapped substantially on most tasks, yet \ours{} had higher mean margins on all eight. Separation was most apparent in text-to-structure generation, with additional rightward shifts in forward reaction prediction and domain/motif prediction; mean differences were smaller on the remaining tasks.

At \ph{1.7B}, higher mean margins again occurred on all eight tasks, with clearer separation across a broader range of tasks. Pronounced shifts appeared in forward reaction prediction, retrosynthesis, property prediction, catalytic activity prediction and domain/motif prediction.

Whereas exact-match accuracy evaluates freely generated answers, these margins measure reference-token preference under teacher forcing. Their agreement in direction shows that improved final-answer accuracy accompanies stronger average preference for reference-answer tokens across heterogeneous scientific tasks, providing an output-level complement to the benchmark results.

\section{Discussion}
\label{sec:discussion}

Our results identify the organization of computation across model depth as a design dimension for scientific foundation models. \ours{} retains the pretrained Transformer block structure but replaces each attention and feed-forward branch's single update with a numerically integrated continuous-depth transformation. Its improvements across eight AI4Science tasks suggest that adapting pretrained representations involves not only scale, data or fine-tuning objectives, but also how transformations are computed within each block.

First, additional within-block computation complements parameter scaling. \ours{} outperforms its parameter-matched Qwen3 counterpart at both scales; at \ph{1.7B}, it matches or exceeds Qwen3-\ph{4B} on seven tasks and exceeds Qwen3-\ph{8B} on three. It also approaches the latter's aggregate accuracy with approximately 15\% less estimated inference computation. These comparisons illustrate why additional solver evaluations are part of the design rather than incidental overhead: they reuse existing parameters to recover part of the performance otherwise obtained by increasing model size.

Second, the recurrent control shows why the organization of repeated computation matters. With the same Qwen3-\ph{0.6B} initialization, base-training corpus and downstream configuration, \ours{} exceeded the Ouro-style control by 5.79 percentage points on average and performed better on six tasks. This supports the integration-based architecture over the evaluated recurrent alternative and makes computation structure, rather than parameter count alone, an explicit design choice.

Third, parameter inheritance makes this adaptation practical. The layer-wise diagnostic identified minimum-loss scales near the original pretrained values for most attention and feed-forward configurations. This supports retaining the inherited scale at initialization, while continued pretraining adapts the parameters to numerical integration. The framework can therefore build on an existing language model rather than learn the backbone from scratch.

Fourth, both numerical integration and adapter capacity shape
adaptation. Two-step RK2 attained the lowest reported loss
at the end of the 20,000-step solver ablations at both
model scales. At \ph{1.7B}, it also showed a steeper fitted
excess WikiText-103 test-loss decline than one-step RK2.
Because both methods are second order, this comparison concerns integration resolution rather than higher formal order. Likewise, increasing LoRA rank from 8 to 16 improved three of four tasks at both scales, whereas rank 32 produced task-dependent gains and losses. These findings support two-step midpoint RK2 with rank-16 LoRA as a practical configuration, without treating maximum solver order or adapter capacity as automatically preferable.

Finally, higher reference-token margins complement the sequence-level gains. Under identical reference prefixes, \ours{} assigns stronger average preference to reference tokens, with clearer distributional separation across more tasks at \ph{1.7B}. Together with exact-match accuracy on freely generated answers, this indicates improvements in both conditional token preference and final-answer correctness, connecting the architectural changes to observable output behaviour.

Several limitations define the scope of these conclusions. The experiments cover Qwen3-derived models at two scales and a block-wise RK implementation; other model families, attention mechanisms and adaptive solvers remain to be examined. Broader generalization also requires additional data sources and scientifically meaningful distribution shifts, including molecular-scaffold and protein-family splits. Finally, reference-based metrics may miss valid alternative scientific outputs, while teacher-forced margins do not capture free-running error propagation. Evaluations that accommodate alternative answers and prospective scientific validation would further establish practical scope.

Overall, \ours{} shows that pretrained language models can be adapted by reorganizing hidden-state computation. Combining pretrained parameter reuse, structured within-block integration and task-specific LoRA offers a complementary route to parameter scaling. Treating hidden-state propagation as an explicit design variable therefore provides a practical direction for more capable and parameter-efficient scientific foundation models.

\section{Methods}
\label{sec:methods}

\subsection{Block-wise continuous-depth conversion}
\label{subsec:methods_conversion}

We initialize our models from pretrained decoder-only language
models in the Qwen3 family~\cite{yang2025qwen3}. We use $j$ to
index residual sublayers and $\ell$ to index Transformer blocks,
each of which contains an attention and a feed-forward residual
sublayer. Let $h_j$ denote the hidden representation entering
residual sublayer $j$. In the original model, the sublayer applies
\begin{equation}
    h_{j+1}=h_j+F_{\theta_j}(h_j),
\end{equation}
where $F_{\theta_j}$ denotes either the self-attention or
feed-forward residual branch, including the corresponding
pre-normalization operation~\cite{vaswani2017attention,he2016deep}.

We replace each discrete residual update with the numerical
solution of a local initial-value problem. For a residual sublayer
with parameters $\theta_j$, the hidden representation is defined
over a normalized local depth coordinate $t\in[0,1]$ as
\begin{equation}
    \frac{\mathrm{d}h(t)}{\mathrm{d}t}
    =f_{\theta_j}(h(t)),\qquad h(0)=h_j.
    \label{eq:local_ode}
\end{equation}
The output of the numerical solver at $t=1$ replaces the original
residual endpoint.

Each converted sublayer inherits the parameters of the corresponding
pretrained residual branch. The tokenizer, token embeddings,
positional encoding, normalization operators and language-model
head are unchanged. The conversion therefore preserves the
representation dimension and parameter initialization while
changing the computation used to propagate representations
through each residual sublayer.

Importantly, Equation~\ref{eq:local_ode} defines a separate local
dynamical system for each attention and feed-forward sublayer.
Parameters are shared across numerical stages within a sublayer
but not across Transformer blocks. The resulting architecture
is therefore piecewise continuous in depth and should not be
interpreted as a single globally parameter-shared neural
ordinary differential equation.

\subsection{Numerical integration of hidden-state dynamics}
\label{subsec:methods_solver}

We convert the self-attention and feed-forward residual branches
separately and apply the corresponding numerical updates sequentially
within each Transformer block. We evaluate three fixed-step solvers:
one-step midpoint RK2, one-step classical RK4 and two-step midpoint
RK2. Unless otherwise stated, \ours{} uses two-step midpoint RK2,
adopted on the basis of the continued-pretraining solver
comparisons described below.

For a generic residual branch with parameters $\theta$, the
dynamics are autonomous: $f_{\theta}$ does not receive the
continuous-depth coordinate $t$ as an explicit input.
The model uses neither time embeddings nor depth-conditioned
parameters. The coordinate $t$ specifies only the integration
interval and numerical step size.

For two-step midpoint RK2, the interval $[0,1]$ is divided into two
uniform steps of size $\Delta t=1/2$. Starting from
$h^{[0]}=h_{\mathrm{in}}$, step $s\in\{1,2\}$ is evaluated as
\begin{align}
    k_{1}^{[s]}
    &=
    f_{\theta}\!\left(h^{[s-1]}\right),\\
    k_{2}^{[s]}
    &=
    f_{\theta}\!\left(
        h^{[s-1]}+\frac{\Delta t}{2}k_{1}^{[s]}
    \right),\\
    h^{[s]}
    &=
    h^{[s-1]}+\Delta t\,k_{2}^{[s]}.
\end{align}
The numerical endpoint is
\begin{equation}
    h_{\mathrm{out}}=h^{[2]}.
\end{equation}
All four function evaluations use the same pretrained parameters, but
each evaluation recomputes the normalization and residual-branch
outputs from its stage-specific hidden representation.

For Transformer block $\ell$, we define
\begin{equation}
    f_{a,\ell}(h)
    =
    \mathrm{SelfAttn}_{\ell}
    \bigl(\mathrm{Norm}_{a,\ell}(h)\bigr)
\end{equation}
and
\begin{equation}
    f_{m,\ell}(h)
    =
    \mathrm{MLP}_{\ell}
    \bigl(\mathrm{Norm}_{m,\ell}(h)\bigr).
\end{equation}
Let $\Phi_{\mathrm{RK2}\text{-}2}(f,h)$ denote the endpoint obtained
by integrating the update function $f$ from the initial state $h$
over $[0,1]$ using two midpoint-RK2 steps of size $\Delta t=1/2$.
Given block input $h^{(0)}$, the self-attention branch is integrated
first:
\begin{equation}
    h^{(1)}
    =
    \Phi_{\mathrm{RK2}\text{-}2}
    \bigl(f_{a,\ell},\,h^{(0)}\bigr),
\end{equation}
followed by the feed-forward branch:
\begin{equation}
    h^{(2)}
    =
    \Phi_{\mathrm{RK2}\text{-}2}
    \bigl(f_{m,\ell},\,h^{(1)}\bigr).
\end{equation}
Here, $h^{(1)}$ is the hidden state after attention integration,
and $h^{(2)}$ is the final block output.

The attention vector field includes the complete self-attention
residual branch, comprising the query, key, value and output
projections. These projections are recomputed at every RK stage using
the stage-specific hidden representation. The causal mask and
sequence-position indices are held fixed across stages, and the
backbone positional encoding is applied using the original token
positions. The feed-forward vector field analogously contains the
complete gated MLP residual branch.

This computation uses repeated evaluations of the same
residual function at intermediate states within a normalized
local interval, rather than introducing independently
parameterized copies of the branch.
Two-step midpoint RK2 uses four function evaluations
per converted residual branch, whereas one-step RK2
uses two, one-step RK4 uses four and the original
discrete residual branch uses one.
We report parameter counts, residual-function evaluations
and estimated FLOPs to characterize the computational
requirements of the evaluated models.

Unless otherwise stated, solver comparisons use the same source checkpoint, training examples, token order, optimizer settings and downstream adaptation protocol. This design controls for differences in initialization and data exposure, while compute-normalized analyses separately account for the different numbers of function evaluations.

\subsection{Layer-wise scale-sweep diagnostic}
\label{subsec:methods_scale}

We use a layer-wise scale sweep to assess whether the original
pretrained parameter scale remains suitable after RK2-based
continuous-depth conversion. The diagnostic is performed on
Qwen3-\ph{0.6B} and Qwen3-\ph{1.7B}, examining attention and
feed-forward residual branches across layers.

For each source checkpoint and layer--branch configuration
$(\ell,r)$, where $r\in\{a,m\}$ denotes the attention or feed-forward
branch, respectively, we apply RK2-based conversion and uniformly
rescale the pretrained parameters under examination:
\begin{equation}
    \theta_{\mathrm{sweep}}^{(c)}
    =
    c\,\theta_{\mathrm{sweep}}.
\end{equation}
Here, $\theta_{\mathrm{sweep}}=\theta_{\ell}$ denotes the
pretrained parameter tensors of the selected Transformer
block $\ell$. Only this block is rescaled by $c$; all parameters
outside it remain at their original pretrained values.
The coefficient $c=1$ preserves the selected block's pretrained
parameter values. Each layer--branch configuration and scaling
coefficient is evaluated independently from the original parameter
values, without accumulating scaling across coefficients or
blocks. No gradient-based parameter updates are performed
during the sweep.


We evaluate each configuration on 1,000 examples randomly sampled
from the training corpus using seed 42. Evaluation uses teacher forcing: the model receives the reference prefix and predicts the next token in the original text. We compute next-token cross-entropy over all valid target-token positions in the sampled texts, excluding padding tokens.
The loss is averaged over the total number of valid target tokens across the 1,000 examples.

Let $M_{\ell,r}(c)$ denote the resulting model configuration.
For each layer and branch type, we record the coefficient
minimizing the teacher-forced language-modelling loss:
\begin{equation}
    c_{\ell,r}^{\star}
    =
    \operatorname*{arg\,min}_{c\in\mathcal{C}}
    \mathcal{L}_{\mathrm{TF}}
    \bigl(M_{\ell,r}(c);\mathcal{D}_{\mathrm{cal}}\bigr),
\end{equation}
where $\mathcal{D}_{\mathrm{cal}}$ denotes the sampled training
examples, $\mathcal{L}_{\mathrm{TF}}$ denotes the teacher-forced
next-token cross-entropy loss, and $\mathcal{C}=\{0.5,0.6,\ldots,1.5\}$ is the grid of
scaling coefficients evaluated during the sweep.

Figure~\ref{fig:base_validation}a shows the loss across coefficients
for each layer and branch type and marks the corresponding
minimum-loss coefficient. Concentration of the optima near $c=1$
supports retaining the original pretrained parameter scale when initializing continuous-depth base training.

\subsection{Continued pretraining after conversion}
\label{subsec:methods_continued_pretraining}

Following conversion, we continue pretraining to adapt the
inherited parameters to the modified numerical computation.
The main models are initialized from Qwen3-\ph{0.6B} and
Qwen3-\ph{1.7B}, respectively, and both use two-step midpoint RK2.
At both scales, continued pretraining uses the same sampled
SlimPajama-627B\footnote{\url{https://www.modelscope.cn/datasets/cerebras/SlimPajama-627B}} corpus. We retain the original Qwen3 tokenizer and pack documents into sequences of 4,096 tokens. Documents are packed without inserting additional end-of-document tokens, and attention is restricted within document boundaries. Padding tokens are excluded from the count of actual training tokens.

Training is conducted on six NVIDIA A100 GPUs using DeepSpeed
ZeRO-2~\cite{rasley2020deepspeed}. All model parameters are optimized
with AdamW using a peak learning rate of $5\times10^{-5}$,
$\beta_1=0.9$, $\beta_2=0.95$ and weight decay
$0.01$~\cite{loshchilov2017fixing}. The learning rate is warmed up
for 1,000 optimization steps and subsequently follows a cosine
decay schedule. Training uses BF16 arithmetic and a global batch
size of 48 sequences. We train for 100,000 optimization steps,
corresponding to a nominal budget of
$100{,}000\times48\times4{,}096=19.6608\times10^9$
token positions, or approximately 20 billion before excluding
padding.

For solver ablations, we train one-step RK2, one-step RK4 and
two-step midpoint RK2 for 20,000 optimization steps at both
the \ph{0.6B} and \ph{1.7B} scales. Within each scale, all
configurations are initialized from the same Qwen3 checkpoint
and use identical training data, batch sizes and optimizer
settings. This ablation budget is separate from the
100,000-step continued-pretraining budget used for the main
models. The training curves in
Fig.~\ref{fig:base_validation}b,c show these 20,000-step
ablation runs. Comparisons at the same optimization step
therefore match training exposure, while the number of
residual-function evaluations varies with the solver.

We use the final checkpoint obtained at the end of continued
pretraining for downstream LoRA adaptation. Training loss is
monitored throughout optimization. At the evaluated checkpoints,
we compute next-token cross-entropy on the official test split
of WikiText-103~\cite{merity2016wiki}.

For the compute-scaling analysis, we use analytically
estimated cumulative training FLOPs. 
Specifically, we estimate cumulative training computation using the
parameter--token approximation
$C_{\mathrm{train}}\approx6PD$~\cite{kaplan2020scaling},
where $P$ is the nominal model parameter count and $D$
is the cumulative number of training token presentations.
To account for repeated residual evaluations, we use
the NFE-adjusted approximation
\begin{equation}
    C_{\mathrm{train},q}
    \approx
    6\nu_q P D,
    \label{eq:training_flop_estimate}
\end{equation}
where $\nu_q$ is the number of function evaluations
per residual branch: two for one-step RK2 and four
for either one-step RK4 or two-step midpoint RK2.
The NFE multiplier is applied at the model level as
a leading-order approximation, rather than through
an operator-by-operator FLOP count.
The WikiText-103 test losses recorded at the evaluated checkpoints are fitted separately for each solver configuration to Eq.~\ref{eq:compute_loss_scaling}.
The asymptotic loss $E_q$, scaling coefficient $A_q$ and exponent $\alpha_q$ are jointly estimated rather than fixing $E_q$ in advance.
We report the fitted exponents and coefficients of determination, $R^2$.
The plotted quantity $L-E$ denotes test loss minus the asymptotic loss fitted for that configuration.

\subsection{AI4Science tasks and datasets}
\label{subsec:methods_benchmark}

We evaluate the models on eight scientific prediction tasks grouped
into four families. The molecule-oriented family comprises reagent
prediction, forward reaction prediction, retrosynthesis and molecular
property prediction~\cite{andronov2023reagent,schwaller2019molecular,
yang2019molecular,wu2018moleculenet}. The protein-oriented family
comprises catalytic-activity prediction and protein domain or motif
prediction~\cite{ryu2019deepec,yu2023enzyme,mistry2021pfam,
blum2025interpro}. The biomolecular-text family contains
chemical--disease relation extraction~\cite{li2016bc5cdr,
wei2016biocreativecdr}. The materials-science family contains
text-to-structure generation~\cite{flamshepherd2023language,
antunes2024crystallm,jain2013materialsproject}.

Seven tasks are derived from Mol-Instructions: reagent prediction,
forward reaction prediction, retrosynthesis, property prediction,
catalytic-activity prediction, domain or motif prediction, and
chemical--disease relation extraction~\cite{fang2024mol}.
Mol-Instructions is a diverse biomolecular instruction dataset
spanning molecule-oriented, protein-oriented and biomolecular-text
tasks. For materials text-to-structure generation, we use the public
\texttt{text2struc\_dataqa} dataset hosted on Hugging Face.
\footnote{\url{https://huggingface.co/datasets/vbv373/text2struc_dataqa}}

The data are divided into training, validation and test partitions in
a ratio of 48:1:1. The same task partitions are used across all
compared models. The eight tasks cover molecular strings, numerical
targets, protein labels or spans, biomedical relation triples and
structured materials outputs. This diversity provides a common
setting for evaluating adaptation across heterogeneous scientific
task formats.

\subsection{Primary and task-specific evaluation metrics}
\label{subsec:methods_metrics}

The primary endpoint in Figure~\ref{fig:ai4sci_results}a is
task-level exact-match accuracy. Predictions are normalized using
task-specific deterministic procedures defined before evaluation.
These procedures include molecular canonicalization, whitespace and
case normalization for constrained textual labels, and schema
normalization for structured outputs. The normalization and
failure-handling rules are applied identically to all models.

For Figure~\ref{fig:ai4sci_results}b, we compute the unweighted
macro-average
\begin{equation}
    S_{\mathrm{macro}}
    =
    \frac{1}{8}\sum_{j=1}^{8}S_j,
\end{equation}
where $S_j$ is the exact-match accuracy for task $j$. This statistic
gives each task equal weight irrespective of dataset size and is
reported together with the eight individual scores.

We additionally report five task-specific metrics per task for the
radar analysis in Figure~\ref{fig:ai4sci_results}c. For reagent
prediction, these are molecular validity, reagent-set exact match,
reagent-level F1, Morgan-fingerprint Tanimoto similarity and
MACCS-fingerprint Tanimoto similarity~\cite{landrum2013rdkit,
schneider2015get,durant2002reoptimization}. Forward reaction
prediction uses molecular validity, product exact match, normalized
Levenshtein similarity, Morgan-fingerprint similarity and
MACCS-fingerprint similarity~\cite{yujian2007normalized}. Retrosynthesis
uses molecular validity, reactant-set exact match, reactant-level F1,
Morgan-fingerprint similarity and MACCS-fingerprint similarity.
Molecular strings are canonicalized before comparison, and invalid
predictions receive a score of zero for chemistry-dependent metrics.

Property prediction is evaluated using MAE, RMSE, Pearson correlation,
Spearman correlation and $R^2$. MAE and RMSE are converted into
positive scores for radar visualization, so that lower prediction
errors correspond to higher plotted values.

Catalytic-activity prediction is evaluated using balanced accuracy,
macro-F1, AUROC, area under the precision--recall curve and Matthews
correlation coefficient~\cite{huang2005using,chicco2020advantages}.
Metrics with range $[-1,1]$, including correlation coefficients and
Matthews correlation coefficient, are mapped to $[0,1]$ for
visualization using $(x+1)/2$.

Domain or motif prediction uses token-level F1, segment-level F1,
boundary intersection-over-union, macro-F1 and exact span match.
Chemical--disease relation extraction uses entity-pair F1,
relation-level F1, triple-level exact F1, macro-F1 and triple recall.
Text-to-structure generation uses schema validity, exact structure
match, field-level F1, numerical-tolerance accuracy and constraint
validity. Radar values are normalized to $[0,1]$ and oriented so that
higher values indicate better performance. The radar plots summarize
complementary aspects of scientific output quality within each task.
For radar visualization, each metric is oriented so that
higher values indicate better performance and then
min--max scaled across the models displayed in the
corresponding task panel. Models omitted from the panel
do not contribute to the scaling bounds. Axis annotations
indicate the metric ranges before radial scaling.

\subsection{LoRA supervised fine-tuning}
\label{subsec:methods_lora}

After continued pretraining, models are adapted separately to each
downstream task using low-rank adaptation
(LoRA)~\cite{hu2022lora}. The final continued-pretraining checkpoint
is used to initialize downstream adaptation. LoRA adapters are
inserted into the query, key, value and output projections of
self-attention and the up, gate and down projections of the
feed-forward branch. The pretrained backbone weights are kept fixed
while the low-rank adapters are optimized on task-specific
instruction data.

For the main Mol-Instructions experiments, LoRA rank is $r=16$ and
the scaling parameter is $\alpha=32$. Optimization uses AdamW with
a learning rate of $2\times10^{-4}$ and zero weight decay. Training
is conducted on six GPUs with DeepSpeed ZeRO-2, BF16 arithmetic, a
global batch size of 192 sequences and a maximum context length of
1,024 tokens. We use cosine learning-rate decay, a warmup ratio of
3\% and train for 20 epochs.

Each task uses identical prompts, target serialization and loss
masking across models. At inference, the models use deterministic
greedy decoding. Prompt formatting, post-processing and evaluation
code are shared across models. Continuous and discrete models are
adapted using the same task data and optimization settings.

We evaluate LoRA ranks 8, 16 and 32 on four representative
AI4Science tasks: retrosynthesis, catalytic activity
prediction, chemical--disease relation extraction and
text-to-structure generation. The comparisons are performed
for both \ours{}-\ph{0.6B} and \ours{}-\ph{1.7B}
to assess the sensitivity of downstream performance to
adapter capacity. The rank-16 setting is the default used in the main
benchmark experiments, and its task-level results
are reused in the rank-ablation comparison.

\subsection{Ground-truth-token margin analysis}
\label{subsec:methods_gt_margin}

We characterize output-level changes using ground-truth-token logit
margins on held-out test examples. Given prompt $x$ and reference
answer $y=(y_1,\ldots,y_T)$ of length $T$, the model is conditioned
at answer position $t$ on the prompt and the reference prefix
$y_{<t}$. The analysis therefore uses teacher forcing with identical
reference prefixes for the compared models.

Let $z_{t,v}$ denote the pre-softmax logit for vocabulary item $v$ at
answer position $t$. We define the ground-truth-token margin as
\begin{equation}
    m_t
    =
    z_{t,y_t}
    -
    \max_{v\neq y_t}z_{t,v}.
\end{equation}
A positive margin indicates that the reference token receives a
larger logit than every competing token; a larger value indicates
greater separation from the strongest competing token.

Margins are computed only at non-padding answer positions and are
summarized separately for each task and model scale. We compare
Qwen3 and \ours{} at the \ph{0.6B} and \ph{1.7B} scales using
the same prompt--answer examples. Density curves show the margin
distributions, and dashed vertical lines indicate the corresponding
means. This output-level analysis measures preference for reference
scientific answers under teacher forcing and complements the
sequence-level exact-match evaluation of freely generated answers.

\subsection{Baselines and comparison protocol}
\label{subsec:methods_baselines}

The baseline set includes Qwen3 models with \ph{0.6B}, \ph{1.7B},
\ph{4B} and \ph{8B} parameters~\cite{yang2025qwen3}, together with
Ouro-\ph{1.4B}~\cite{zhu2025scaling}. We additionally construct
Qwen3-\ph{0.6B}-Ouro, an Ouro-style recurrent control initialized
from the same Qwen3-\ph{0.6B} checkpoint as \ours{}-\ph{0.6B}.
The looped control uses the same base-training corpus and downstream adaptation configuration as the continuous model. Both the recurrent control and \ours{}-\ph{0.6B} use four evaluations of each original attention and feed-forward residual function, matching the branch-level function-evaluation budget.

We distinguish three comparison types:
\begin{enumerate}
    \item \emph{Same-scale comparison}: \ours{} versus Qwen3
    models at the same parameter scale under a common downstream
    LoRA adaptation protocol;
    \item \emph{Recurrent-control comparison}: \ours{}-\ph{0.6B}
    versus Qwen3-\ph{0.6B}-Ouro under a common pretrained
    initialization, base-training corpus and downstream adaptation
    configuration;
    \item \emph{Resource comparison}: predictive performance
    compared across model parameter scales and estimated inference
    FLOPs under the standardized evaluation protocol.
\end{enumerate}

The same-scale comparisons evaluate adaptation at a fixed model
scale. The recurrent control compares alternative organizations of
repeated computation, while larger Qwen3 models and Ouro-\ph{1.4B}
provide reference points for the parameter--performance and
compute--performance comparisons.

All internally adapted models use the same task splits, prompts,
decoding settings and evaluation scripts. Parameter counts and
estimated inference FLOPs are used to characterize the corresponding
resource--performance trade-offs.

\subsection{Ablation design}
\label{subsec:methods_ablation}

The ablation experiments are designed to distinguish four possible
sources of performance differences. 
First, the layer-wise scale sweep evaluates the suitability of the inherited parameter scale after RK2 conversion using teacher-forced language-modelling loss on 1,000 randomly sampled training examples.
Second, solver ablations compare one-step RK2, one-step RK4 and two-step RK2 under both token-matched and compute-aware settings. 
Third, the Qwen3-\ph{0.6B}-Ouro control compares continuous-depth integration with an Ouro-style recurrent alternative under a common pretrained initialization, base-training corpus and downstream adaptation configuration.
Fourth, LoRA-rank ablations test sensitivity to downstream adapter
capacity.

These experiments characterize pretrained parameter-scale suitability, solver-dependent optimization, compute--loss scaling, recurrent computation and downstream adaptation capacity.





\subsection{Statistical analysis and reproducibility}
\label{subsec:methods_reproducibility}

Main downstream benchmark results and the recurrent-control
comparison are averaged over three independent LoRA
fine-tuning runs. For each model, the base checkpoint is
held fixed across these runs, while the LoRA fine-tuning
seed is varied under the same task splits, data and
optimization settings. The repetitions are performed only
at the downstream adaptation stage and do not involve
retraining the base model.

Task-level results are reported separately, and the aggregate
benchmark score is the unweighted mean exact-match accuracy
across the eight tasks. Aggregate scores are computed within
each run before averaging across runs.
Model comparisons use the same test examples, prompt
templates and evaluation procedures.
Task-specific metric values are mapped to $[0,1]$ for
radar visualization, with all axes oriented so that higher
values indicate better performance. The aggregate benchmark
score is computed from the original exact-match accuracies,
not from the normalized radar values.

Code, configuration files and evaluation scripts will be
released upon acceptance to support reproduction of model
conversion, continued pretraining, downstream adaptation
and evaluation. The release will include training
configurations, preprocessing scripts, prompt templates
and evaluation code used for the reported figures and
benchmark comparisons.

\section{Data availability}

The datasets analysed in this study are publicly available. The molecule-oriented, protein-oriented and biomolecular-text tasks are derived from the Mol-Instructions dataset hosted on Hugging Face at \url{https://huggingface.co/datasets/zjunlp/Mol-Instructions}. Mol-Instructions is distributed under the CC BY 4.0 license. The materials text-to-structure task is based on the public dataset hosted on Hugging Face at \url{https://huggingface.co/datasets/vbv373/text2struc_dataqa}. No internal or proprietary datasets were used in this study. The processed evaluation files, task splits and benchmark metadata used to reproduce the reported results will be released together with the code upon acceptance.

\section{Code availability}

Code for model conversion, continuous base training, downstream LoRA fine-tuning, evaluation, metric computation and figure generation will be released upon acceptance. The release will include configuration files and scripts sufficient to reproduce the main benchmark results, parameter--performance analysis, mechanistic validation, ablation studies and interpretability analysis reported in this paper.

\section{Acknowledgements}

This work was supported by the National Natural Science
Foundation of China (Grant No.~62441617); the Postdoctoral
Fellowship Program and the China Postdoctoral Science
Foundation (Grant Nos.~2024M764093 and BX20250485);
the Beijing Natural Science Foundation
(Grant No.~4254100); and the Beijing Advanced Innovation
Center for Future Blockchain and Privacy Computing.

\section{Author contributions}

Zhaoxin Fan led the project, supervised the research, secured the research resources, and participated in revising the manuscript. Zheng Zhu provided part of the computational resources. Yuqing Ma independently verified the data splits, evaluation metrics, and figure inputs, and contributed to the analysis and correction of the issues identified during verification. Faguo Wu substantively reviewed the continuous-depth parameterization, the RK solver design, and the Methods section, and revised the mathematical formulations and method exposition. Zhiming Zheng participated in discussions on the research positioning, scientific significance, and the boundaries of the conclusions, and substantively revised the corresponding discussions. Jian Yang, under Zhaoxin Fan's supervision, designed and performed the study: he developed the continuous-depth parameterization and the RK-based solver, implemented the code, carried out the experiments and analysis, and wrote the initial draft. Wenjun Wu reviewed the experimental design and baseline settings, participated in the interpretation of the results, and contributed to the revision of the manuscript. All authors read and approved the final manuscript.

\section{Competing interests}

The authors declare no competing interests.

\bibliography{sn-bibliography}

\end{document}